# Deep neural networks architectures from the perspective of manifold learning

German Magai
*Department of Big Data and Information Retrieval*
*Higher School of Economics*
Moscow, Russia
gmagaj@hse.ru

*Abstract* — Despite significant advances in the field of deep learning in applications to various areas, an explanation of the learning process of neural network models remains an important open question. The purpose of this paper is a comprehensive comparison and description of neural network architectures in terms of geometry and topology. We focus on the internal representation of neural networks and on the dynamics of changes in the topology and geometry of a data manifold on different layers. In this paper, we use the concepts of topological data analysis (TDA) and persistent homological fractal dimension. We present a wide range of experiments with various datasets and configurations of convolutional neural network (CNNs) architectures and Transformers in CV and NLP tasks. Our work is a contribution to the development of the important field of explainable and interpretable AI within the framework of geometrical deep learning.

*Keywords* — Representation learning, Deep networks architectures, manifold learning, performance evaluation.

## I. INTRODUCTION

Over the years, deep neural networks (DNNs) have shown success in applications in many areas: natural and social sciences, computer vision, text, audio processing and generation, and more. Despite the obvious empirical progress, some deep learning questions are still poorly understood theoretically: why some architectures are more successful than others at solving the same problem? What determines the generalizing ability of DNNs and how can it be assessed without using a test sample? A promising and important area of research is the interpretation of the learning process and the internal representation of DNNs in terms of topology and geometry.

Recently, many efficient DNNs architectures have been developed. Transformer neural network architectures are increasingly used not only in the field of language processing, but also in other tasks that can be represented in terms of sequence processing. For example, Vision Transformers [56] in image processing tasks, where Convolutional Neural Network (CNN) have been used until recently. And there is still no clear answer which approach to neural networks' design is better. Analysis and comparison of different architectures from different aspects: internal representation, robustness, performance, scaling etc. can greatly improve understanding of why some architectures perform better than others. New results in this direction will help provide approaches to the design and development of efficient architectures in various tasks.

In this paper, we analyze knowledge representation in neural networks of different architecture families. To interpret the internal representations of deep models, we use an approach based on manifold learning and topological data analysis (TDA). We touch upon the issue of the evolution of the embedding space throughout the depth of neural networks using the example of various architectures, activation functions and datasets. Furthermore, we propose a method for estimating the generalization ability based on the topological and geometric properties of the internal representation. We pioneered the use of fractal dimension to analyze the internal representation of DNNs. Experiments include CNNs, Vision Transformers and large language models. As a result of the empirical study, our contributions and conclusions are as follows:

• In the process of data propagation throughout the depth of the DNN, the properties of data representations change. In our work, we show a significant difference in the evolution of data in DNN belonging to different families of architectures: based on convolutions and attention mechanisms.

• During the training of the DNN, the geometric and topological properties of the embedding space evolve, the model tries to learn the highest quality representations. We propose a way to estimate the expected performance and generalization ability of models based on representation analysis without using the standard test-training split.

## II. RELATED WORK

Approaches based on topology and geometry can give a new vision of the above problems related to understanding deep learning algorithms and finding their original, effective solution by formulating a descriptive model of learning processes and the structure of neural networks. Also relevant for our study are works that systematically study and compare the features of different architectures according to various criteria. The available results can be divided into 3 groups: topological and geometric aspects of learning and performance of DNNs, research on the internal representation of DNNs, and comparative analysis of neural networks. Some articles aim to clarify the processes that take place inside a DNN during training and inference, while others suggest improvements to existing DL solutions.

Particular attention is paid to the analysis of the generalizability, capacity and expressiveness of deep learning models using topology and geometry methods. Gus in [2] raises the question of whether the learning and generalizability of a DNNs depends on the homological complexity of a particular dataset. On the other hand, authors of [3] and [4] investigate the performance of DNNs depending on the ID of the training dataset and notice that the generalization error does not depend on the extrinsic dimension of data. In [5] it is proved that a DNNs can successfully learn with help of SGD to solve a binary classification problem when a number of conditions on the width, length and sample complexity are met with certain geometric properties of the training dataset. There is also a series of works that consider the expressive power of DNNs through the analysis of the topology of the decision boundary [6-8]. In some works, a computational graph of a neural network is considered [9-10]. It was proved [77] that the Hausdorff dimension of the SGD trajectory can be related to generalization error. [78] consider the $PH_{dim}$ of the learning trajectory in the parameter space, in the optimization SGD process, this is the first work that connects $PH_{dim}$ and theoretical problems of deep learning. [11] analyzing the space of weights in convolution filters, the authors come to the conclusion that the topological features of this space change during the learning of CNN.

To understand the learning process of DNNs, it is important to have an idea of the dynamics and changes in geometric properties within the space of embeddings on different layers. The paper [17] argues that the learning process of DNNs is associated with the untangling of object manifolds, on which the data lie, when passing through the layers of the DNN, and a quantitative assessment of the untangling is proposed. The authors of [18] analyze the change in the geometry of object manifolds (dimension, radius, capacity) during training and argue that as a result of the evolution of object manifolds in a well-trained DNN, by the end of the hierarchy of layers, manifolds become linearly separable. [19] focus on the problem of memorization and generalization when training DNN and come to the conclusion that memorization occurs on deeper layers due to a decrease in the radius and size of object manifolds. [20] also consider transformations of data in a DNN and attempts to formulate deep learning theory in terms of Riemannian geometry. In [67] the ability of DNN to learn an efficient low-dimensional representation is considered, and [68] deals with the DNN-based transformation of data into minimal representation. Paper [79] uses persistence landscapes to analyze the dynamics of topological complexity across all layers of the neural network, according to their results, topological descriptors are not always simplified in the learning process.

In many ways, our work is a continuation and improvement of the ideas outlined in [21], where it is argued that when passing through the layers of a FC neural network, the data topology is simplified and the Betti numbers are reduced, and the ReLU activation function contributes to better performance. But the methodological problem of this work is that it is necessary to calculate the Betti numbers for a fixed scale parameter. Our work uses TDA and removes this need for epsilon search. In addition, we make use of the results of [22], where shows the dynamics of ID when passing data on different layers throughout the depth of the model and its relationship with performance and generalization gap using the example of several modern CNNs architectures, the authors conclude that neural networks transform data into low-dimensional representations. Paper [23] shown that clusters with different density peaks are formed on different layers, which reflects the semantic hierarchy of the ImageNet. In [12] mean-field-theory is used to analyze the geometric properties of data in the internal representation of the BERT language model. Work [13] compares different DNN architectures from decision boundary perspective. In [65] the problem of calibrating modern architectures into classification task is explored. Paper [14] also shows that CNNs and ViT «see» the data representation in different ways. But in our work, we come to the conclusion that models of different architectures transform the geometric and topological properties of data in different ways, which is confirmed experimentally.

III. PRELIMINARIES

We can consider various aspects of deep neural networks: decision boundaries, loss function landscape, functional graph, weight dynamics during optimization. But it is also very important to consider internal representations, analyze their dynamics at different epochs of learning and at depths. It is possible to analyze data properties from different points of view, including using methods of applied algebraic topology and geometry. In this section, we present a definition and a brief description of the topological data analysis and manifold learning concepts that we will use to explore data properties within a DNN.

*A. Deep neural networks*

A deep neural network is a function DNN: $R^x \rightarrow R^y$ defined by the composition DNN = softmax$\circ f_d \circ ... \circ f_2 \circ f_1$, where $f_i$ - is a function of the i-th structural block, softmax – output activation function, y - number of classes, x - dimension of data and $M^x{}_i = f_i(M^n{}_{i-1})$ object manifold in embedding space generated as a result of the action of block $f_i$, and the composition of functions is denoted by the $\circ$ symbol.

At different depths, CNNs extract features corresponding to different levels of abstraction; on the first layers, features correspond to the image style and general geometric primitives, and at the end – to high level specific features that separate some classes from others. Vision Transformer (ViT) is based on the attention mechanism, the input image is divided into patches and encoded, thus the image is represented as a sequence of tokens. The attention mechanism is used to quantify pairwise interactions between tokens in a sequence. The design of large language models (LLM), such as BERT (Bidirectional Encoder Representations from Transformers), RoBERTa and GPT-2, is also based on the attention mechanism, which was originally invented specifically for NLP tasks. It is thanks to attention that LLM began to show impressive results in the problems of classification and generation of sequences.

In the process of propagation through layers throughout the depth of CNN, ViT or BERT, data is transformed and representations are disentangled. But different architectures transform data in different ways, as will be shown in our work. The problem of classification in deep learning is reduced to the decomposition of linearly inseparable object manifolds into compressed, compact representations that can be separated by a linear hyperplane [18, 70]. This assumption is applicable to both neural networks and brain models, and is also consistent with the findings of neuroscience [70]. And the mechanism of the Transformer is similar to the principle of the hippocampus of the brain [82].

## B. Persistent homology

The main tool for topological data analysis is the concept of persistent homology. This construction summarizes the process of changing the Betti numbers during the filtration. Details and more deep theoretical introduction can be found in [11, 83, 24 - 26]. To be able to work with objects in vector space X, it is necessary to construct simplicial complexes K from them. We use the Vietoris-Rips complex $K(X; r)$, $\{i_0,...,i_s\}$ is a simplex of $K(X; r)$ if balls of radius r around $x_{i_0},...,x_{i_s}$ intersect pairwisely. For a simplicial complex K, $H_i(K)$ denotes the vector space of i-th simplicial homology of K (coefficient field F is less important for us, hence omitted). $\beta_i(K) = \dim H_i(K)$ denotes the i-th Betti number. For i = 0, 1, 2 this number counts the number of connected components, cycles, and cavities in K respectively. An increasing sequence of simplicial complexes is called a filtration: $\{K_i\}_{(i \in \mathbb{Z} \geqslant 0)} = K_0 \subseteq K_1 \subseteq K_2 \subseteq ... \subseteq K_s$.

Persistent homology is an invariant of a filtration, which tracks the change in homology throughout the filtration. The persistent homology is encoded by the barcode: that is a finite collection of triples $(t_{birth}^n, t_{death}^n, \deg^n)$, where $t_{birth}^n$ is the birth time of n-th homological feature, $t_{death}^n$ its death time, and $\deg^n$ its degree (Fig. 1). We restrict to degrees 0, 1 in our work due to the resource-intensive calculation of higher degree. $l^n = t_{birth}^n - t_{death}^n$ the lifetime of the n-th homology feature, called lifespan, it is obvious that it cannot be negative.

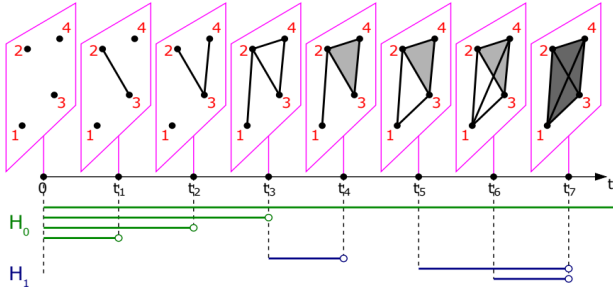

Fig. 1. An example of a barcode with discrete time $t_i$, green color denotes homology of dimension 0, blue – 1.

A long interval in barcode means that we have a sufficiently long-lived n-homology, which is a persistent topological feature, it helps to evaluate the topological properties of a space. An equivalent representation of the barcode is the persistent diagram. This is a plane with coordinates of the birth $t_{birth}$ and death $t_{death}$ times, each homology is indicated by a point with a color corresponding to the degree. The power-weighted sum of N lifespans for the i-th homology degrees is denoted as follows

$$E_\alpha^i(X) = \sum_{n=1}^{N} l_n^\alpha \qquad (1)$$

where $\alpha \geqslant 0$, it is noted in [27] that $E_1^0(X_n)$ it is equal to the length of Euclidean minimal spanning tree (MST) of set of n points in $\mathbb{R}^d$. $E_\alpha^1(X_n)$ can be understood as the cyclomatic capacity of the data, the larger this value, the more obvious the cycles. Finding a geometric interpretation of $E_\alpha^1(X_n)$ could be a topic for future research. $E_\alpha^i(X_n)$ we will call topological descriptors and use them to analyze the properties of data $X_n$ in our work. Numerical experiments and calculations of persistent diagrams were carried out using Ripser library [28].

## C. Manifold learning and fractal dimension

We are increasingly faced with very high-dimensional data in computer vision problems, sound signal processing, in the analysis of gene expression, and others. According to the manifold hypothesis [30], the data X lies on a low-dimensional submanifold: $X \subseteq M^n \subseteq \mathbb{R}^d$, where d – extrinsic dimension, n – ID. [31] Proved that given sufficient sample complexity, we can model a low-dimensional representation of the data with some error. It is possible to divide the methods for estimating the ID into two classes [32, 33]: 1) Local methods estimate the dimension at each data point from its local neighborhood, and then calculate the average over the local estimates of the ID: NN algorithm [34], MLE [35], nonlinear manifold learning methods. 2) Global methods estimate the dimension using the entire dataset, assuming that the dataset has the same dimension throughout: PCA, MDS, k-NNG, GMST [36]. This group also includes fractal-based methods: Hausdorff dimension, correlation dimension and box-counting dimension.

If we consider the persistent diagrams obtained from a set of points in $\mathbb{R}^d$, it is often possible to notice an accumulation of homological features in the diagonal area and near zero, they are born and die quickly. These points are considered to be topological noise that does not carry useful information. However, the decay rate of «noise» depends on the size of the submanifold on which the set of points lies. [39] introduces the concept of Persistent homological fractal dimension $PH_{dim}$, generalizing [38] for any homology group dimensions and based on [76]. X - bounded subset of a metric space and μ a measure defined on X, for each $i \in \mathbb{N}$ define the $PH_{dim}$ of μ:

$$PHdim_\alpha^i(\mu) = \frac{\alpha}{1-\beta} \qquad (2)$$

where

$$\beta = \limsup_{n \to \infty} \frac{\log\left(\mathbb{E}\left(E_\alpha^i(x_1,..x_n)\right)\right)}{\log(n)} \qquad (3)$$

and $x_1,..,x_n$ are sampled independently from μ. That is, $PHdim_\alpha^i(\mu) = d$ if $E_\alpha^i(x_1,..x_n)$ scales as $n^{\frac{d-\alpha}{d}}$ and $\alpha \geqslant 0$. Larger values of α give relatively more weight to large intervals than to small ones. In other words, the persistent homological dimension can be estimated by analyzing the asymptotical behavior at $n \to \infty$ of $E_\alpha^i(x_1,..x_n)$ for any i. [39] proved that if μ satisfies the hypothesis of Ahlfors regularity, then $PHdim_\alpha^0(\mu)$ equals the Hausdorff dimension of the support of μ.

In practice, the $PH_{dim}^i\alpha(\mu) = d$ can be calculated as follows: for real $m_1, m_2,...,m_k$ logarithmic values $\log_{10}(n) = m_i$, we randomly sample n points from X and calculate the $E_\alpha^i(x_1,..x_n)$. We then use linear regression to fitting the power law for obtained values n and $E_\alpha^i(x_1,..x_n)$. In all experiments in our work, the calculations were carried out for the case i = 0 and α = 1, and the $PHdim_1^0(\mu)$ will be denoted $PH_{dim}$. We restrict ourselves to i = 0, because this will allow us to get all the comprehensive geometric information. Semantically, different $PH_{dim}$ is expressed in the diversity. The larger the ID, the more diverse the data (Fig 2).

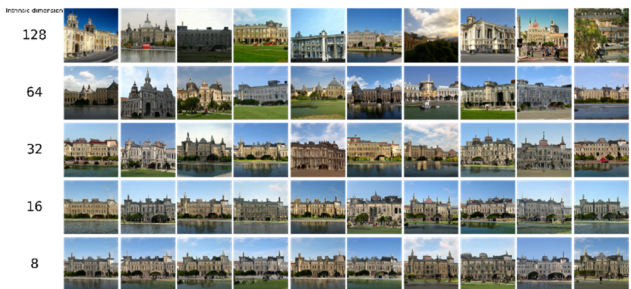

Fig 2. Relationship between ID and diversity of the Castles image dataset, rows correspond to different ID

TABLE I. ESTIMATION OF THE SYNTHETIC DATASETS ID

| ID\real ID | 8 | 16 | 24 | 32 | 48 | 64 | 96 |
|---|---|---|---|---|---|---|---|
| $PHdim_0$ | 8.96 | 17.72 | 23.49 | 30 | 32 | 43.2 | 46.58 |
| TwoNN | 10 | 20 | 24 | 28 | 30 | 38 | 36 |
| MLE | 11 | 19 | 23 | 26 | 28 | 33 | 28 |
| CorrDim | 10.17 | 12.97 | 16.47 | 17.34 | 18.4 | 15.8 | 14.05 |

Unfortunately, the problem of existing algorithmic methods for calculating the ID is that for a correct estimate, an exponentially large amount of data is needed, which leads to a systematic error and a significant underestimation of the obtained ID [33]. According to [4] we can control the upper bound ID of the dataset: generate synthetic data using a BigGAN, to set the necessary upper bound d for the ID, we fix 128 - d elements of the hidden vector equal to zero. We generate synthetic datasets with different ID for assessing the accuracy of the $PH_{dim}$ method and comparing with other approaches. The generated examples with didderent ID are shown in Fig. 2. In table 1 we compare the accuracy of $PH_{dim}$ with other ID estimation approaches: Correlation dimension (CorrDim) [74], TwoNN [34], Maximum likelihood estimation (MLE) [35]. As you can see, $PH_{dim}$ estimates the ID more accurately than others. The datasets from Table 1 are synthetic data generated with BigGAN by controlling ID.

## IV. EVOLUTION OF TOPOLOGICAL DESCRIPTORS AND FRACTAL DIMENSION IN DEEP MODELS

To develop new, high-performance deep networks architectures and solve open problems in the field of deep learning, it is necessary to understand the essence of the processes that occur during training, as well as what affects the expressive power and ability of deep learning models to generalization. DNN learns an internal low-dimensional representation of the data manifold on different layers and deep learning models training can be interpreted in terms of evolution of embedding manifold representations, changes in topology and geometry of data. An embedding manifold $X_n \subseteq R^d$ is a low-dimensional representation of the object data manifold within a DNN, where d - the width of the layer. In the process of passing the data manifold from from block to block along the all depth, the emdedding manifold's characteristics change. Here we track the change in topological descriptors and $PH_{dim}$ at different depths of model and associate them with the success of training in image and text classification tasks. Based on our experiments, we test the hypothesis that, in terms of topology and internal representation geometry, deep neural networks of different architectures transform data in different ways. We are clarifying exactly how the architecture affects data dynamics.

### A. Experimental setup

The experiments include a systematic comparison of DNNs in terms of analyzing the dynamics of the embedding space in supervised classification regime. We consider several families of architectures:

• Deep networks based on the convolution mechanism: ResNet [41], SE-ResNet [42], MobileNetV2 [43], VGG-18 [44].

• VisionTransformer based on the attention mechanism that work with images as with sequences of tokens.

• ConvMixer [57] model is a hybrid of the architectural features of CNN and ViT, it combines the advantages of both approaches to architecture design. It abandons the classic pyramidal structure of CNN and uses image partitioning, as in ViT.

• Large attention-based language model (LLM) BERT [58], have replaced the generation of RNN language models. BERT consists 12 blocks, each encoder layer uses bidirectional attention, which allows the model to consider context from both sides of the token. RoBERTa [81] is an improvement on BERT.

Datasets in image classification task: CIFAR-10 [45], Street View House Numbers [46], ImageNet [47]. For the task in the NLP domain, we use different emotional coloring tweets dataset [80]. The tweets dataset includes 8 basic emotions: anger, anticipation, disgust, fear, joy, sadness, surprise, and trust. For correct comparison of models with each other, all CNN consist of 10 structural blocks, and 128 channels, in order to exclude the influence of the ambient space. ViT also consists of 10 blocks. GlobalAveragePooling operations are used to vectorize feature maps in CNN, and first sequence token for Transformers. Each model was trained with an accuracy of 95% or more on the training dataset using the Adam [53] optimizer with adaptive learning rate. Train/test split 0.8/0.2, hardware specifications: Nvidia GeForce GTX 3080Ti, Intel Core i7-10700K, 16 GB of RAM.

### B. Experiments

The design ideology of modern deep learning architectures is based on the principle of structural blocks (moduls). The architecture consists of specific blocks connected in series, which include a set of different operations, such as activation, 1D, 2D and depthwise convolution, attention, pooling, skip-connections, compression, batchnormalization, etc. Different configurations and combinations of these operations give different efficiency, accuracy, and computational complexity and at the same time affect differently the change in the geometry of the object manifold.

We analyze the embedding representation at the output of each structural block in CNNs, ViT, ConvMixer and LLMs. One of the main results of this section is the demonstration of the behavior of the topological descriptors and $PH_{dim}$ of the data embedding manifold on different blocks during training on CIFAR-10 and text dataset at different epochs. In Fig. 3, 4, 5, each line indicates the dynamics of the topology and geometry of the data over the entire depth, line has its own color from bright yellow to blue, which indicates the accuracy on the test dataset at this stage of training. For experiments, we feed a randomly selected batches of 300 examples from train dataset of each class to the input of the DNNs several times, and then average the results across batches and over all 10 classes.

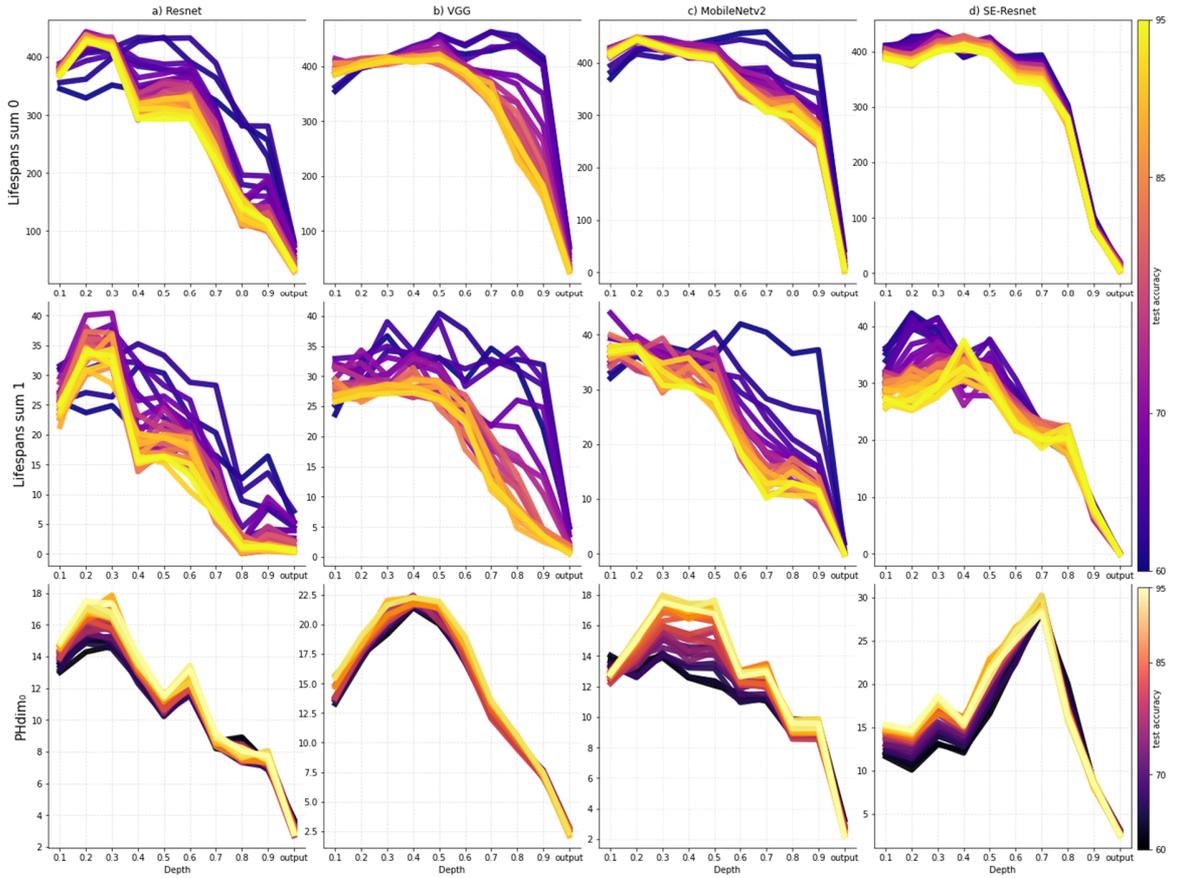

Fig. 3. Change in topological descriptors and PH$_{dim}$ (y-axis) on different blocks of CNNs during training at different epochs, x-axis – relative depth from 0.1 (output of 1 block) to 1 (last output fully connected layer). Columns denote various CNN architectures, rows denote topological and geometric properties. Dataset – CIFAR-10

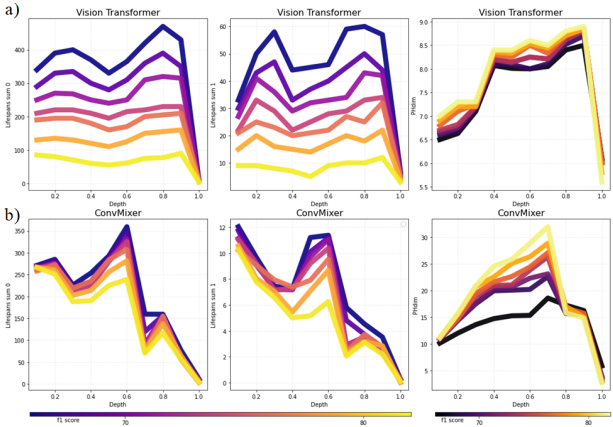

Fig. 4. Changing the topological descriptors and PH$_{dim}$ inside ViT and ConvMixer across all depth at different epochs. Dataset CIFAR-10

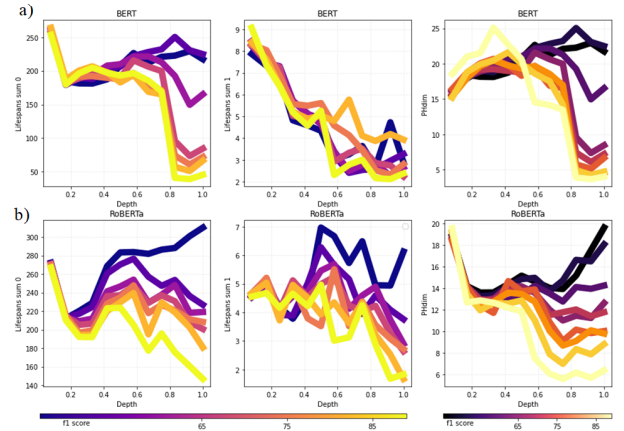

Fig. 5. Changing the topological descriptors and PH$_{dim}$ inside BERT and RoBERTa across all depth (12 encoder layers) at different epochs.

According to the observation (Fig.3) of the evolution of the topology of a train dataset manifold when passing through a CNNs, in the process of training a DNN to learn more efficiently and faster to lower the values of topological descriptors at the all depth. On different architectures of CNN, the evolution of the data manifolds occurs in different ways, this may be a consequence of the unique arrangement of structural modules (blocks) that underlie a particular architecture.

In the case of a ResNet, the dynamics of a decrease in topological descriptors shows a decrease from the very first blocks, and in the case of MobileNetV2 with linear bottleneck and SE-ResNet with squeeze-and-excitation block, the topology is simplified from the middle of the layer hierarchy. Experiments show that the dynamics of topological and geometric characteristics of models based on the mechanism of convolutions and attention is different.

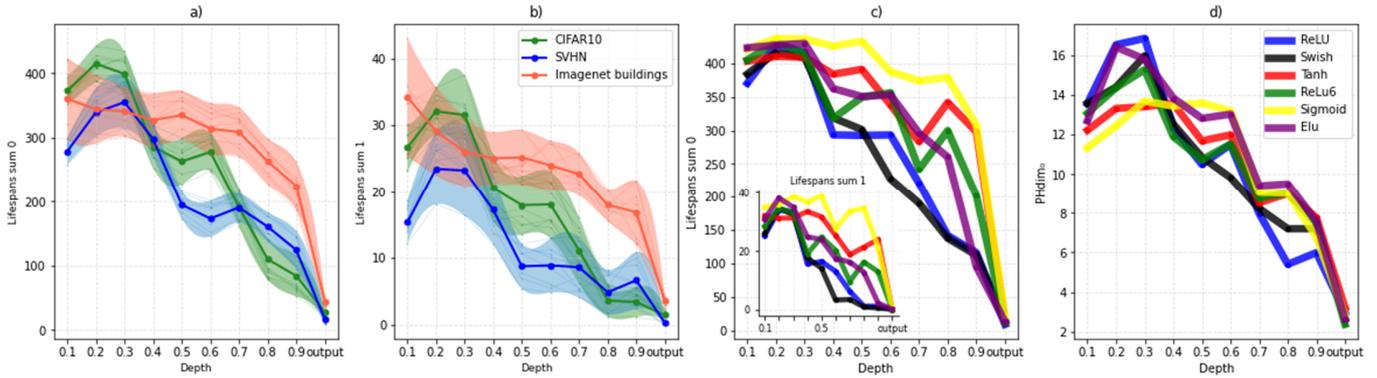

Fig. 6. Changing the topological descriptors and $PH_{dim}$ inside ResNet: a, b) Different datasets, thin lines indicate different classes. c) Influence of activation functions on changing topological descriptors. d): Influence of activation functions on changing $PH_{dim}$.

A decrease in topological descriptors $E_1^0(X_n)$ indicates that the classes in the dataset $X_n$ are becoming more compact. $E_1^1(X_n)$ indicates that the number of noticeable cycles in the data is decreasing, which also indicates about compactification and structural simplification. While we observe a decrease in topological descriptors over the entire depth in CNN, in ViT the values of these descriptors almost do not change, however, depending on the epoch of their training, the indicators are smaller (Fig 4). In the case of $PH_{dim}$, empirical observations also show a different picture, $PH_{dim}$ increases at intermediate layers in ViT. It is important to note that a sharp decrease in $PH_{dim}$ on last output fully connected layer is expected and is explained by the small dimension of the ambient space. But in CNN, we see the hump shape as the data evolves throughout the depth. The decrease in topological features in the CNN can be associated with the pyramidal hierarchical structure of the CNN, in which the information obtained at the first ones is generalized in deeper layers. In ViT there is no hierarchical structure and image resolution does not change with depth, all layers have access to the same knowledge representation. ConvMix is a hybrid of CNN and ViT, the dynamics of changing topological and geometric properties in the embedding space is more similar to CNN (Fig 4.b).

In the embeddings space of large language models BERT (Fig. 5a) and RoBERTa (Fig. 5b), geometric and topological properties change throughout the depth, but in a different way. The topological descriptors of not well trained models do not decrease over the entire depth, but rather increase. The trajectory of the $PH_{dim}$, unlike CNN and ViT, does not reproduce the shape of the hump so clearly. This observation suggests that in the internal representation, large language models «understand» and transform data in a different way.

In the process of training on different layers, the data form clear clusters that correspond to their classes. As shown in this section, the internal presentation of data is simplified across the layers, if one measures simplicity by natural topological descriptors. We can propose the following explanation for the phenomenon observed in Fig. 3,4,5: In an ideal well-trained DNN the internal representation at the last layer should represent a collection of untangled and separated clusters, ideally a finite set of points. A finite set of points has $PH_{dim} = 0$. So it is natural to expect that $PH_{dim}$ gets reduced throughout the layers, which is consistent with observations in next chapter.

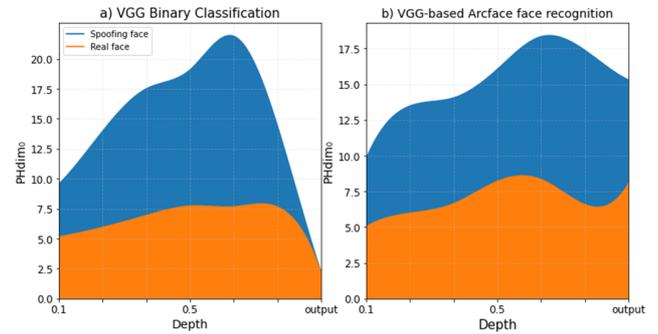

Fig. 7. The dynamics of $PH_{dim}$ face anti-spoofing manifold. a) Binary classification. b) Arcface face recognition.

For all datasets within the ResNet representation, there is the same tendency to simplify topological descriptors, but in different ways. On fig. 6a, 6b show the difference in data dynamics in the same architecture but with different datasets: CIFAR-10 [45], 10 classes 32×32×3 images, Street View House Numbers (SVHN) and 10 classes corresponding to "buildings" from ImageNet 64×64×3. The trend towards simplification is observed for all classes in datasets; in Fig. 6a, 6b, all classes are indicated by thin lines of different colors. As shown in Fig. 6a, b the topological properties for the SVHN dataset are simplified faster, indicating that this is an easier task for ResNet.

We evaluate the change in topological descriptors and $PH_{dim}$ on intermediate ResNet-101 layers' representation with different activation functions. As you can see in the Fig. 6c, 6d, different activation functions make different contribution to the topology and $PH_{dim}$ change; with the Swish activation function (black line), model has better accuracy on the test dataset in our experiments, and according to [48] Swish improves CNN performance over ReLU (blue line) and Sigmoid (yellow line), helping to alleviate the vanishing gradient problem. The ReLU, ReLu-6 and ELU activation functions are more successful in practice than Tanh and Sigmoid, which is consistent with the dynamics of topological simplification of the train dataset manifold in our experiments. Success of DNN with ReLU is explained by they are not homeomorphic map [21].

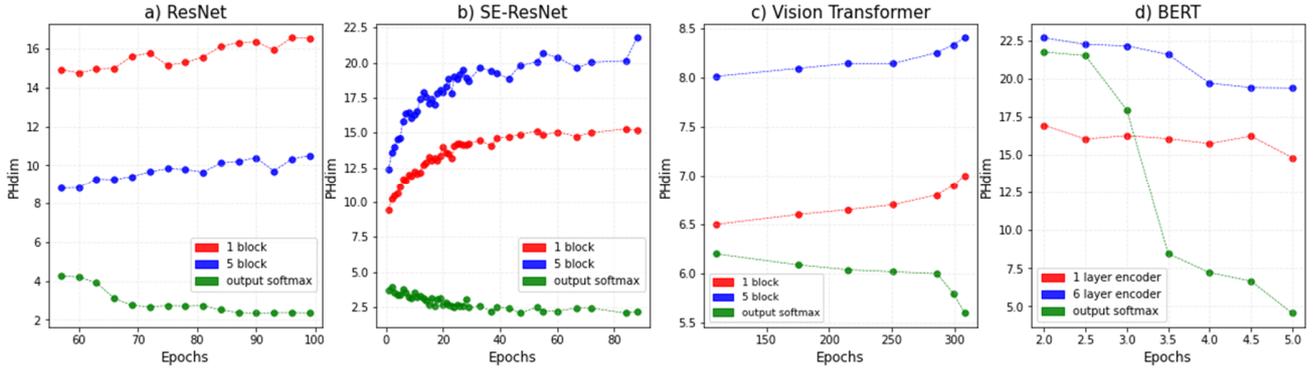

Fig. 8. PH$_{dim}$ on first, middle and last layers for all training epochs of ResNet, SE-ResNet, ViT and BERT models.

We analyze how different datasets are transformed in the internal representation of neural networks in terms of PH$_{dim}$, Fig. 7 explans on the experiments in Fig. 6. Real face and spoofing examples face were chosen as datasets. Spoofing examples are a type of attack on face recognition systems where a physical photo or screen of a photo is shown instead of a person's real face [54].

We show here that the PH$_{dim}$ of the real and spoofing faces dataset [59] in the internal representation of the DNNs changes as it passing through all the layers (Fig 7). In the binary classification problem (spoof/real, 2000 examples per each class), the dynamics of the face data manifold repeats the shape, as in the Fig. 3. In the problem of 10 persons face recognition, the VGG [44] with fixed 32 width architecture is used with a loss function ArcFace [62], in the output space the objects lie on a 32-dimensional hypersphere, which simplifies the comparison of objects of the same class with each other. PH$_{dim}$ in this case does not correspond to the shape of the hump. Examples of spoofing attacks have a higher PHdim than real faces, due to the presence of effects related to glare, texture, lighting and other effects, the visual complexity of spoofing attacks is higher. An estimate of the PH$_{dim}$ can be useful to determine the quality and diversity of a spoofing examples dataset. Tests have shown that the CelebA-Spoof [60] dataset PH$_{dim}$ = 14, while the screen attack dataset from [59] PH$_{dim}$ = 10. This can be explained by that CelebA-Spoof presents a wider range of spoof attack types, not just screen attacks.

### C. Discussion and comparison with other works

Here we will discuss how the results in this chapter relate to existing research about comparing different architectures. The gradual increase in the PH$_{dim}$ in CNN up to the middle of the depth can be explained by the pyramidal structure of CNN models. Low-level features are extracted in the first layers, and high-level features are extracted in the intermediate layers closer to the middle. Semantically more complex feature maps with more information are formed, which explains the increase of PH$_{dim}$. This correlates with [64], where an analysis of "visual complexity" of feature maps on different layers is proposed and it is shown that visual complexity is higher in the middle layers than in the first and last layers. One can interpret the decrease in PH$_{dim}$ on the last layers as the model tends to learn a simpler and more efficient representation of the extracted features on the previous layers.

The dynamics of feature maps on the surface and deep layers at different stages of learning does not differ significantly for CNN and ViT (Fig. 8a, 8c). We see that CNN models recognize visually and semantically more complex feature maps that carry more information, so we can observe an increase in ID on these layers. ViT do not have a hierarchical structure, but at the same time, they themselves extract more and more semantically complex features in the learning process. In BERT, at different learning epochs, PH$_{dim}$ decreases at all intermediate layers, not just at the last one (Fig. 8d). This indicates a fundamentally different logic of knowledge processing in the fields of image and text processing.

The paper [14] presents a systematic comparison of CNN and ViT architectures in terms of centered kernel alignment (CKA) similarity. Many results of our work confirm the mentioned emirical study, while we use geometry and topology methods to analyze representations. We notice the following similarities: the representation structure of ViT and CNN from the CKA similarity point of view shows significant differences: ViT have very similar representations throughout the depth of the model, while ResNet models show much less similarity between the lower and deeper layers. This correlates with our observations. The higher the ID of the embeddings in the intermediate layers, the more "qualitatively" the deep model is trained, but the higher the ID in the output layer, the less the test accuracy of the model. In the case of ViT and CNN in deep and shallow layers, PH$_{dim}$ increases with each training epoch. This may indicate that intermediate layers in CNN and ViT are trying to learn more complex visual representations with more and more information with each epoch.

The conclusion of the chapter is that we tested the hypothesis that the data transformation in the internal representation of different architectures is very different. The statement that the model gradually reduces topological properties is not true for all types of architectures; The statement is true for CNN and BERT. But this does not apply to ViT - at all depth levels, the topology almost does not change. Also, not very well trained LLM do not reduce topological properties while CNNs do, although not gradually. We proposed a geometric interpretation of the phenomenon of changing the semantic complexity of feature maps in CNN, in the learning process, the fractal dimension on intermediate layers, except for the last ones, increases, also for architectures based on the attention mechanism (ViT), we observe similar behavior.

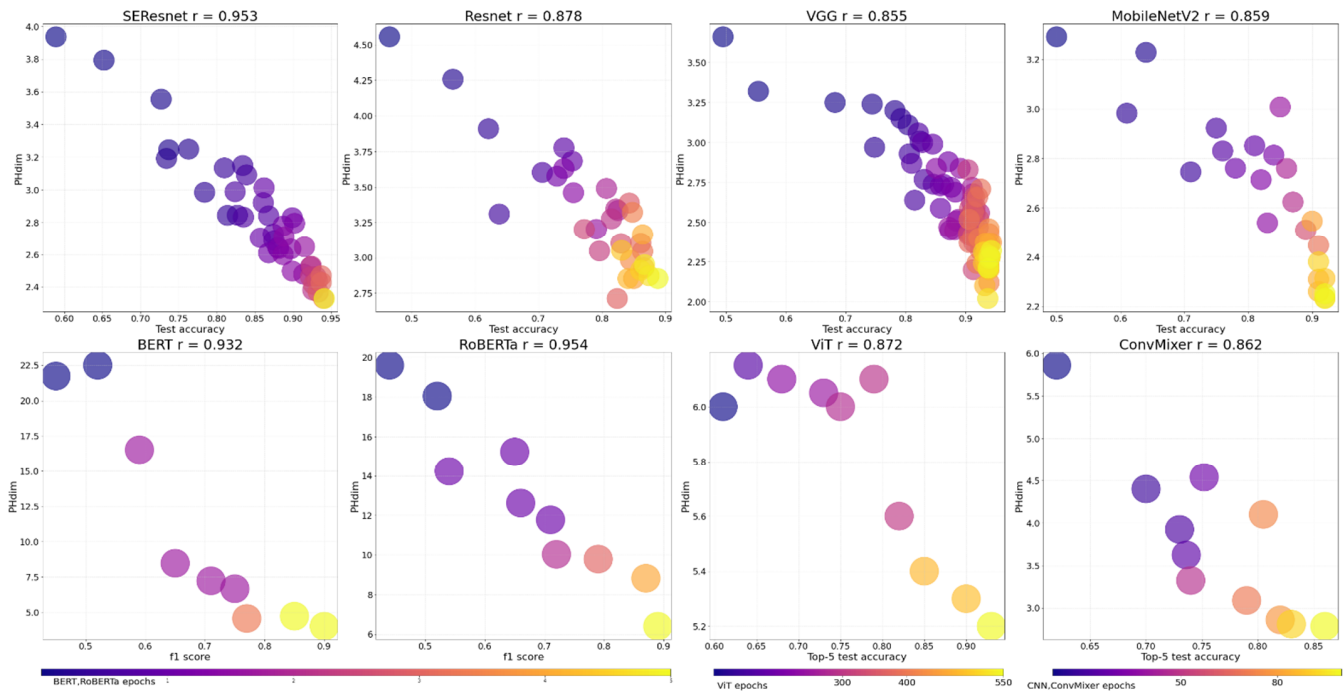

Fig. 9. Dependence between PH$_{dim}$, obtained from the internal representation of the last layer output and the test accuracy at different epochs. Pearson's correlation coefficient denoted by r.

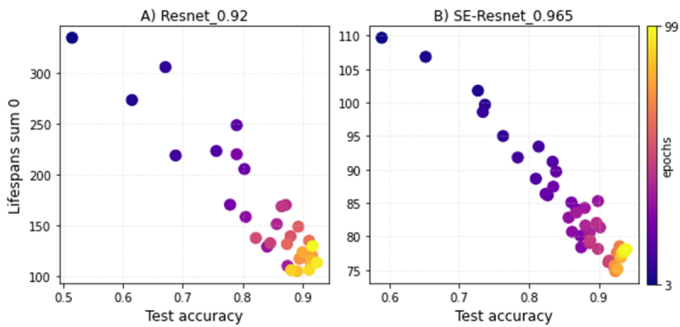

Fig. 10: Dependence between $E_0^1(X)$, obtained from the iternal representation X of the last layer and the accuracy on the test set in the learning process at different epochs. a) ResNet, r = -0.92. b) SE-ResNet, r = -0.965.

## V. INFLUENCE OF TOPOLOGY AND GEOMETRY ON THE GENERALIZATION ABILITY OF DEEP MODELS

Neural network generalizations are the ability to learn from the training set to work effectively with data that was not in the training dataset. The measure of generalization in classification problems is determined by the concept of a generalization gap - the difference in the accuracy metric between the indicators on the test and train datasets.

In addition to classical measures of DNNs expressivity, such as VC-dimension [66] or Rademacher complexity [74], there are many approaches to estimating generalization error [49 - 52]. [9, 22, 72, 78] demonstrate methods for analyzing the performance and expressive power of DNN based on geometric and topological characteristics.

In this section, we test the hypothesis that neural network training is associated with a gradual simplification of topological descriptors and a decrease of fractal dimension in the internal representation al last layer. We propose a way to evaluate the performance and generalizability of CNN and Transformer models in image and text classification problems. We use PH$_{dim}$ to analyze the dynamics of the geometric properties of the neural network output representation during training at different epochs. PH$_{dim}$ was chosen as a more accurate and mathematically sound way to estimate ID than others (Table 1). We also demonstrate the relationship of topological descriptors to the performance of DNNs trained with different hyperparameters.

### A. Evolution of topological descriptors and PH$_{dim}$ in trainig process

If we consider the last output layer after softmax activation as the final representation $R^k$ of the DNN in the k-classification problem, then we can see a statistically significant relationship between the PH$_{dim}$ of the train dataset batch in this representation and the test accuracy (Fig. 9). We can evaluate the performance of models at different epochs without using the standard split test/train, which is important in conditions of limited data.

The fact that PH$_{dim}$ tends to 0 during the learning process can be related to the theory of Neural Collapse (NC) [40, 55], according to which, in the output feature space $R^k$, data clusters (classes) are concentrated around centroids lying at the vertices of the Simplex Equiangular Tight Frame [55]. And PH$_{dim}$ = 0 is the stage when the clusters corresponding to the classes collapse to a point and the DNN seems to be trying to reach this state during the learning process. This is consistent with Fig.8, where it is empirically shown that the DNN tends to learn the most efficient low-dimensional representation on the last layer, and the rate of PH$_{dim}$ decrease affects the learning accuracy.

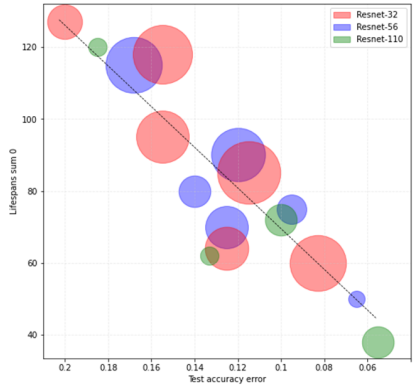

Fig. 11. Generalization gap estimation. The relationship between model accuracy on a test dataset and topological descriptors $E_1^0(X_n)$ in internal representation, ResNet type models, r = -0.9, 50 models. Architectures within one ResNet family form clusters indicated by circles with different colors

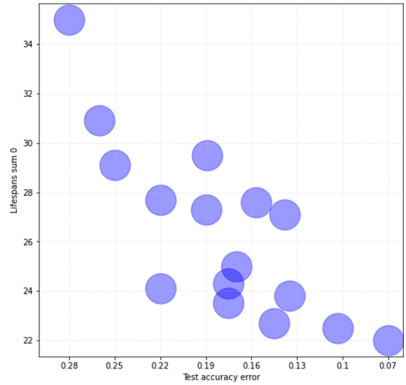

Fig. 12. Generalization gap estimation. The relationship between model accuracy on a test dataset and topological descriptors $E_1^0(X_n)$ in internal representation, Vision Transformer, r = -0.867, 16 models.

As shown in Section 4, the learning of CNN affects the dynamics of topological descriptors throughout the depth. We can see (Fig. 10) the relationship between topology of emdedding space in last intermediate layer and the learning epoch. With each training step and increasing accuracy, the topological descriptors decrease, which confirms the hypothesis that the learning process of a neural network is associated with a gradual simplification of the topology.

*B. Generalization gap estimation*

To estimate the generalization error, it is necessary to consider not one model at different learning epochs, but a family of trained models with different accuracy on the test data. If the learning process can be understood as a simplification of the topology of the data manifold across all depth, then the models with the simplest data on the last layer $X^{out}$ have the best performance in test dataset. Topological descriptors as well as $PH_{dim}$ decrease throughout the depth of the DNN, and they are similarly generalizing ability predictors.

Our hypothesis is supported and a statistically significant inverse correlation r = -0.9 (Pearson's correlation coefficient) can be observed between the topological description $E_1^0(X^{out})$ and the test accuracy error (difference between train accuracy and test accuracy), and this can be a reliable predictor of DNN generalization in the classification problem (Fig 11, 12). To use a topological predictor, only the train dataset is required without the need for a test data, as in Section 5.A.

For experimental verification, a dataset of trained 50 CNN-models of the family of architectures ResNet-32, ResNet-56, ResNet-110 (Fig 11) with different learning hyperparameters was formed: regularization, learning rate, weight decay, batch size, with or without augmentation. All models are trained on the CIFAR-10 dataset to 95-99% train accuracy. To test the approach for assessing the generalizing ability (Fig 12) in the case of other architectures, the Vision Transformer architecture was chosen with a depth of 10 transformer layers, embedding dimension 128, 4 attention heads, patch size 6. And 16 ViT models were trained with different hyperparameters, as well as CNN.

In this section, we have demonstrated that the topological and geometric invariants of the internal representations at the last layer are closely related to the learning process. These invariants can be predictors of the performance and generalizability of deep models across different architectures.

## VI. CONCLUSIONS AND FUTURE WORK

As a result of our empirical study, we have attempted to clarify some aspects of DNNs learning based on the topology and geometry of the internal representation. Several hypotheses were tested and it was shown that neural network architecture significantly influences knowledge representations and data transformation across the depth of the model. Because of the structural features and differences of CNN, ViT and large language models transform data in different ways. How deep neural networks "see" and understand, data can be interpreted geometrically.

We conclude that the learning process of DNNs is related to changes in the topological descriptors and fractal dimensionality of the data manifold. It is shown that the performance and generalization ability of classifiers can be related to the topology and geometry of the data.

Future research may be focused on the analysis of analogies between neuroscience, biological models of the nervous system and artificial neural network models of different architectures, on fundamental similarities and differences. The topic of understanding the decision boundarys of neural networks in terms of combinatorics and polyhedral topology is also very promising. A more theoretical understanding of the observations obtained in this paper may be a topic for future research.

## VII. ACKNOWLEDGMENT